# Nondestructive Chicken Egg Fertility Detection Using CNN-Transfer Learning Algorithms

Shoffan Saifullah[1,2], Rafal Drezewski[1,3], Anton Yudhana[4], Andri Pranolo[5,6], Wilis Kaswijanti[2], Andiko Putro Suryotomo[2], Seno Aji Putra[2], Alin Khaliduzzaman[7], Anton Satria Prabuwono[8], Nathalie Japkowicz[9]

[1]Institute of Computer Science, AGH University of Krakow, Krakow, 30-059, Poland
[2]Department of Informatics, Universitas Pembangunan Nasional Veteran Yogyakarta, Yogyakarta, 55281, Indonesia
[3]Center of Knowledge Engineering and Data Science, Department of Electrical Engineering and Informatics, Universitas Negeri Malang, Malang, 65145, Indonesia
[4]Department of Electrical Engineering, Universitas Ahmad Dahlan, Yogyakarta, 55191, Indonesia
[5]College of Computer and Information, Hohai University, Nanjing 211100 China
[6]Department of Informatics, Faculty of Industrial Technology, Universitas Ahmad Dahlan, Yogyakarta 55166 Indonesia
[7]Faculty of Agricultural Engineering and Technology, Sylhet Agricultural University, Sylhet-3100, Bangladesh
[8]Faculty of Computing and Information Technology in Rabigh, King Abdulaziz University, Jeddah 21589, Saudi Arabia
[9]Department of Computer Science, American University, Washington, DC, 20016, USA

## ARTICLE INFO

## ABSTRACT

This study explored the application of CNN-Transfer Learning for nondestructive chicken egg fertility detection for precise poultry hatchery practices. Four models, VGG16, ResNet50, InceptionNet, and MobileNet, were trained and evaluated on a dataset (200 single egg images) using augmented images (rotation, flip, scale, translation, and reflection). The training results demonstrated that all models achieved high accuracy, indicating their ability to learn and classify chicken eggs' fertility. However, variations in accuracy and performance were observed when these models were evaluated on the testing datasets. The InceptionNet exhibited the best overall performance, accurately classifying fertile and non-fertile eggs. It demonstrated excellent performance in all parameters of the evaluation metrics for both training and testing datasets. When evaluated on the testing datasets, it achieved an accuracy of 0.98, a sensitivity of 1 for detecting fertile eggs, and a specificity of 0.96 for identifying non-fertile eggs. The higher performance is attributed to its unique architecture, efficiently capturing features at different scales, which leads to improved accuracy and robustness. Further optimization and fine-tuning of the models might be necessary to address the limitations in accurately detecting fertile and non-fertile eggs using other models. This study highlighted the potential of CNN-transfer learning for nondestructive fertility detection and emphasized the need for further research to enhance the models' capabilities and to ensure accurate classification.

**Corresponding Author**:

Shoffan Saifullah, Institute of Computer Science, AGH University of Krakow, Al. Mickiewicza 30, 30-059 Kraków, Poland
Department of Informatics, Universitas Pembangunan Nasional Veteran Yogyakarta, Jl. Babarsari 2 Yogyakarta, 55281, Indonesia
Email: saifullah@agh.edu.pl; shoffan@upnyk.ac.id

## 1. INTRODUCTION

Nondestructive and early detection of hatching egg fertility may play a vital role in poultry farming and hatchery practices [1] by determining whether an egg is fertile or infertile without disrupting the incubation process [2]. The early detection and separation of infertile eggs from an incubator may reduce the contamination of incubated eggs. Conventional methods often involve candling and cracking the egg or disturbing the incubation environment, posing risks of reduced hatchability [3]. Thus, researchers have been exploring advanced image processing techniques combined with deep learning algorithms for nondestructive fertility detection to overcome these limitations [4], [5]. However, very limited research is available on applying advanced machine learning tools (e.g., deep learning algorithms) for precise hatchery practices, although Convolutional Neural Networks (CNNs) with transfer learning have emerged as a popular approach for image-based egg fertility detection [6]–[9]. Transfer learning takes advantage of pre-trained CNN models, which have been trained on large-scale datasets like ImageNet, and applies them to specific tasks with limited labeled data [10]. By leveraging the learned knowledge from the extensive dataset, a suitable transfer learning model enables effective feature extraction and may improve the performance of egg fertility detection [11], [12]. Therefore, this research investigated the effectiveness of various CNN-transfer learning algorithms for chicken egg fertility detection. The selected models, including VGG16, ResNet50, InceptionV3, and MobileNet, have been widely recognized for their strong performance in image classification tasks [13]. With their distinct architectures and complexities, these models were evaluated comprehensively to identify the most suitable algorithm for chicken egg fertility detection.

Nondestructive chicken egg fertility detection has been constantly evolved by researchers with various study interests, driven by ongoing research and advancements [14]–[16]. Recent studies have made significant progress by exploring various models and techniques tailored to this domain. These include the utilization of Mask R-CNN for precise segmentation and fertility classification [17], [18], convolutional neural networks [19] for detecting the fertility of multiple eggs simultaneously [15], [20], artificial neural networks for fertility prediction [21], SVM classifiers [22] for extracting first-order statistical features [23], and deep learning models for identifying cage-free hens on a littered floor [24], [25]. These developments demonstrate the potential of deep learning approaches and machine learning algorithms for nondestructive chicken egg fertility detection [26]–[28]. Rapid advancements mark the current state-of-the-art in this field as researchers continually strive to enhance nondestructive chicken egg fertility detection [22]. Adopting deep learning techniques plays a crucial role in creating highly accurate and efficient models for nondestructive chicken egg fertility detection, ultimately contributing to producing healthy chicks for the poultry industry.

The research contribution of this study lies in its systematic and rigorous evaluation of various CNN-transfer learning algorithms for nondestructive chicken egg fertility detection. By thoroughly investigating different models, including VGG16, ResNet50, InceptionV3, and MobileNet, the study provides valuable and deeper insights into their efficacy in accurately classifying fertile and infertile chicken eggs. Furthermore, the research contributes significantly by curating a meticulously acquired dataset comprising 200 labeled images, evenly distributed between 100 fertile and 100 infertile eggs, obtained through the egg candling method and subjected to meticulous image processing techniques. Incorporating sophisticated data augmentation methods, such as random reflections, scaling, rotation, and translation, enhances the model's generalization capability and overall fertility detection performance meticulously. Thus, the study findings may underscore the immense potential of CNN-Transfer Learning for nondestructive fertility detection, holding promising implications for advanced hatchery practices and sustainable production of healthy chicks in the poultry industry in the near future.

This paper was structured to provide a comprehensive overview. Section 1 introduces the background of the study, highlighting the significance of nondestructive chicken egg fertility classification. Section 2 reviews relevant works with related topics and summarizes previous research on chicken egg fertility classification. Section 3 focuses on explaining the detailed CNN-transfer learning methods for chicken egg fertility classification, especially the concept of transfer learning and the adaptation of pre-trained CNN models. Section 4 describes the experimental setup and presents the results obtained from the study, including data collection, preprocessing techniques, model training, and evaluation metrics. Finally, the conclusion summarizes the findings, implications, and potential directions for further improvement and application of nondestructive chicken egg fertility classification using CNN-transfer learning algorithms for the next-generation poultry and hatchery practices.

## 2. RELATED WORKS

Several research works have recently been conducted on nondestructive chicken egg fertility detection, employing various techniques and methodologies [29]–[32]. Researchers have explored the use of computer vision and machine learning algorithms for fertility classification based on visual features extracted from egg images [18], [23], [33]–[35] for this specific purpose. Their proposed approaches achieved promising results, demonstrating the potential of image analysis techniques in accurately determining the fertility status of eggs.

Boynukara *et al.* (2016) studied the fertility discrimination of eggs using ultrasound [36]. The study used a machine vision system to capture ultrasound images of the eggs and analyzed the images to distinguish fertile eggs from infertile ones. The results showed that the system could accurately detect fertile eggs with a high degree of accuracy. Moreover, the study demonstrated the potential for ultrasound imaging as a nondestructive approach for fertility detection in chicken eggs.

In addition to ultrasound imaging, other nondestructive approaches have been explored for fertility detection in chicken eggs. These include dielectric measurements, thermal imaging, and machine vision [37]. Hashemzadeh *et al.* (2017) developed a machine vision algorithm to distinguish fertile eggs from infertile ones using hyperspectral imaging [35], [38]. Deep learning techniques for egg fertility detection using hyperspectral imaging were employed by Çevik *et al.* (2022) in [18]. A nondestructive detection system based on machine vision was designed to identify the fertility of eggs before virus cultivation [12], [39]. These studies demonstrate the promising results for various nondestructive approaches for the fertility detection of chicken eggs.

Ultrasound imaging uses high-frequency sound waves to produce visual images of the internal structures of a body, which is commonly used in medical diagnosis and obstetric ultrasonography [40]. Using ultrasound imaging for nondestructive fertility detection of chicken eggs is promising as an alternative to traditional methods that require the destruction of an egg.

Infrared thermography has also been investigated as a potential nondestructive method for fertility assessment [33]. Some researchers [33], [41]–[43] developed a technique that combined infrared thermography with deep learning algorithms for egg fertility classification. Their approaches exhibited high accuracy in determining the fertility of eggs. However, addressing variations in egg surface temperature and ensuring robustness in different environmental conditions using these approaches is still challenging [44].

Moreover, the previous studies also exerted certain limitations despite the promising results of various nondestructive techniques in detecting the fertility of eggs [22]. Some studies lacked robustness and generalization capability, as they were developed and evaluated on limited datasets [33], [45], [46]. Other studies relied on specialized equipment or complex setups, hindering their practicality and widespread adoption [47]. Additionally, certain studies focused on specific modalities or features without exploring the benefits of integrating multiple techniques [48], [49].

To address these limitations, we conducted the study aiming to leverage the power of transfer learning and deep convolutional neural networks (CNNs) for nondestructive detection of chicken egg fertility. By utilizing pre-trained CNN models and fine-tuning them on a large dataset of annotated egg images, we expected to achieve higher accuracy and robustness in fertility classification. Transfer learning allows models to leverage knowledge learned from large-scale datasets, even when the target dataset is relatively small, enhancing their performance.

Furthermore, to evaluate the performance and generalization capability of the developed models, we employed K-Fold Validation with k=5. This validation technique divided the dataset into k subsets, allowing the model to be trained and evaluated k times with a different subset as the validation set in each iteration. By averaging the performance metrics across the k iterations, we could obtain a more reliable estimate of the models' performance and ability to generalize the unseen data.

By building upon the strengths of previous research and addressing their weaknesses, our approach aims to advance a nondestructive detection of chicken egg fertility, providing a more accurate, robust, and practical solution for the poultry industry.

## 3. MATERIALS AND METHODS

The dataset used in this study contained 200 labeled images [50], with an equal distribution of 100 fertile and 100 infertile chicken eggs [51]. These images were acquired using the egg candling method [52], [53], which allowed visual observation of the embryos inside the eggs [54].

This research aimed to develop a classification model using Convolutional Neural Network (CNN)-Transfer Learning with hyperparameter tuning and optimization algorithms to enhance performance. This proposed approach includes fine-tuning the pre-trained models for detecting the fertility of eggs. The datasets used in this study were divided into training and testing datasets, with a 4:1 ratio. Data augmentation techniques, such as random reflections, scaling, rotation, and translation, were applied to the training datasets to enhance the model's generalization capability. Performance evaluation was conducted based on accuracy, which measures the percentage of correctly classified eggs. This metric provided insights into the effectiveness of each CNN-transfer learning algorithm in detecting fertile and infertile eggs. Therefore, the most suitable algorithm for nondestructive detection of chicken egg fertility could be determined by comparing the performance of different models.

The research workflow consisted of several detailed steps, as illustrated in Fig. 1, which include data preprocessing, image augmentation involving rotations, flips, scaling, translations, reflections, selection of pre-trained CNN models (VGG16, ResNet50, InceptionNet, MobileNet), hyperparameter tuning, model training, testing using a separate dataset, and result analysis. The performance evaluation of the developed models utilized the confusion matrix to assess accuracy, sensitivity, specificity, and precision regarding the fertility classification.

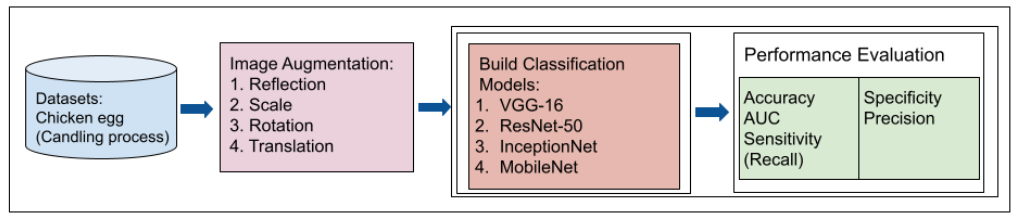

**Fig. 1.** Workflow of the process and performance evaluation of nondestructive chicken egg fertility detection

### 3.1. Dataset

This research utilized datasets of single egg images against a black background [55], [56]. Acquisition of the dataset involved employing the widely recognized candling technique, which entails illuminating eggs with an LED flashlight and capturing low-resolution images using a smartphone camera (13 MP) within a controlled dark room environment [54] (Fig. 2a). Candling is a standard practice extensively employed in the poultry industry to assess egg development and fertility [52].

The dataset encompassed 200 single egg images, each meticulously subjected to rigorous image processing techniques to ensure their suitability for comprehensive analysis. The processing pipeline comprised two essential steps: cropping and segmentation. Extraneous background elements were effectively removed through exact cropping, directing attention solely toward the eggs. Afterward, a precise segmentation procedure was applied to accurately isolate the eggs within the images, enabling focused analysis and subsequent processing. The resulting segmented egg images formed the foundation for ensuing preprocessing [57] and fertility detection procedures.

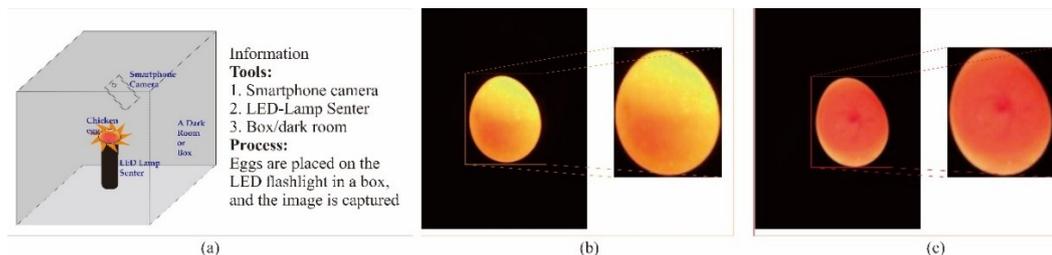

**Fig. 2.** (a) The designated image acquisition system and analyzed results of image acquisition and cropping of (b) fertile (day 7) and (c) infertile chicken eggs

The dataset was thoughtfully divided into two distinct categories to ensure thorough evaluation and adequate training of the employed CNN-transfer learning algorithms [58]. The two categories were fertile and infertile eggs [45], each comprising precisely 100 images. This balanced distribution allows for robust model development, accommodating comprehensive coverage of fertile (Fig. 2b) and infertile eggs (Fig. 2c) during algorithmic training and subsequent assessment stages [12].

### 3.2. Data Augmentation

In the study on nondestructive chicken egg fertility detection, we employed image augmentation techniques to enhance the datasets. Image augmentation involves applying various transformations to existing images, increasing their diversity, and improving the model's generalization capability [59]. In this case, we applied the following augmentation techniques to the segmented egg images obtained from the candling and cropping process (part of Fig. 2b and Fig. 2c).

1. Rotation: We performed random rotations on the segmented egg images within a specified range of angles. By rotating the images, we introduced variations in egg orientations, simulating eggs that appear at different angles during the candling process. This augmentation technique enables the model to learn and recognize eggs from different perspectives.
2. Flip: Horizontal flipping was applied to the images, creating mirrored images of the original dataset. This technique helps address any potential bias caused by the orientation of the egg during the candling process. Flipped images made the model more robust to images of eggs facing different directions.

3. Scale: We employed scaling techniques to resize the segmented egg images. We adjusted the scale factor, both increasing and decreasing, to simulate egg size variations. This augmentation allows the model to learn and adapt to eggs at different distances from the camera, enabling it to detect the fertility of eggs regardless of the size.
4. Translation: We applied random translations to the segmented egg images. Translation involves shifting the image horizontally and vertically within a specific range. By performing translations, we introduced spatial variations in the dataset. This augmentation technique accounts for the possibility of eggs not being perfectly centered during the candling process.
5. Reflection: We used reflection, precisely vertical flipping, to create additional variations of the segmented egg images. This augmentation helps capture eggs that have a flipped orientation or appear differently due to reflections or other factors. The model learns to detect fertility regardless of the egg's apparent orientation by including reflected images.

The specific augmentation techniques used in this study were carefully selected to simulate real-world variations and challenges encountered during egg fertility detection. By incorporating rotations, flips, scaling, translations, and reflections, we expanded the dataset, allowing the CNN-transfer learning algorithms to learn more robust features and improve the performance in detecting egg fertility.

It is essential to know that each augmentation technique's specific range or parameters varies based on the dataset characteristics and the research objectives. We fine-tuned these parameters through experimentation to ensure that the augmented dataset adequately represented the potential variations encountered in real-world scenarios.

### 3.3. CNN-Transfer Learning for Image Classification

In this study, we employed the CNN-Transfer Learning approach to perform image classification for chicken egg fertility detection with the architecture model shown in Fig. 3. Our dataset consisted of only 200 annotated images, which presented a challenge due to the limited data available for training. To overcome this limitation, we leveraged the power of transfer learning and utilized pre-trained CNN models, namely VGG16, ResNet50, InceptionNet, and MobileNet.

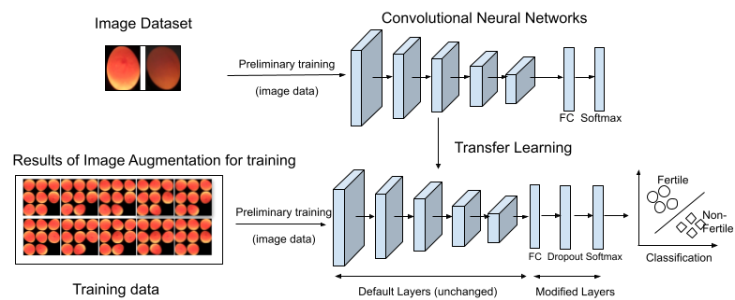

**Fig. 3.** CNN-Transfer-learning architecture model for chicken egg fertility classification

VGG16 is a deep convolutional neural network architecture proposed by the Visual Geometry Group (VGG) at the University of Oxford. It has been widely recognized for its simplicity and effectiveness [60]. VGG16 consists of 16 convolutional layers, followed by three fully connected layers. The architecture includes multiple stacked 3×3 convolutional layers with a stride of 1, padding of 1, and max-pooling layers with a 2×2 window and a stride of 2. This repetitive structure results in 41 layers. The model can capture intricate image features, making it suitable for image classifications. In our proposed method, VGG16 had 134.2 million trainable parameters.

ResNet50 is a deep residual network architecture introduced by Microsoft Research. It was designed to address the problem of vanishing gradients in intense neural networks [61]. ResNet50 consists of 50 layers, including residual blocks that incorporate skip connections. These skip connections allow the network to learn residual mappings, significantly easing the training of deeper networks. The skip connections enable the gradients to flow directly through the network, mitigating the vanishing gradient problem. The ResNet50 architecture has residual blocks with different numbers of layers (e.g., 3, 4, 6, or 9 layers), resulting in a total of 177 layers. In our proposed method, ResNet50 had 23.8 million trainable parameters.

InceptionNet, or GoogLeNet, is a deep convolutional neural network architecture proposed by Google. It introduced the concept of inception modules, which perform efficient feature extraction at multiple scales [62]. The InceptionNet architecture employs 1×1, 3×3, and 5×5 convolutions in parallel to capture features at different levels of abstraction. It also includes pooling operations and concatenating feature maps to capture local and global information. The InceptionNet architecture promotes both depth and width in the network by utilizing multiple branches with different kernel sizes. In our proposed method, InceptionNet had 315 layers and 21.8 million trainable parameters.

MobileNet is a lightweight convolutional neural network architecture for mobile and embedded platforms. It focuses on efficiency and computational simplicity while maintaining reasonable accuracy. MobileNet utilizes depth-wise separable convolutions, which split the standard convolutional operation into depth-wise and point-wise convolutions [63]. This approach significantly reduces the number of parameters and computational complexity compared to traditional convolutional layers. MobileNet balances model size and accuracy, making it suitable for resource-constrained environments. MobileNet had 154 layers and 2.2 million trainable parameters in our proposed method.

In the proposed method, we leveraged these pre-trained architectures (VGG16, ResNet50, InceptionNet, MobileNet) and fine-tuned them using our limited 200 annotated chicken egg images dataset. By fine-tuning, we retained the learned features from the pre-trained models while adapting them to the specific chicken egg fertility classification. This allowed us to utilize these architectures' knowledge and feature extraction capabilities while customizing them specifically for our objectives.

Overall, our proposed method utilized the power of CNN-Transfer Learning and pre-trained models (Fig. 4) to address the challenge of limited data availability in chicken egg fertility classification. The specific architecture details and trainable parameters for each model were as follows:

- VGG16: 41 layers and 134.2 million trainable parameters.
- ResNet50: 177 layers and 23.8 million trainable parameters.
- InceptionNet: 315 layers and 21.8 million trainable parameters.
- MobileNet: 154 layers and 2.2 million trainable parameters.

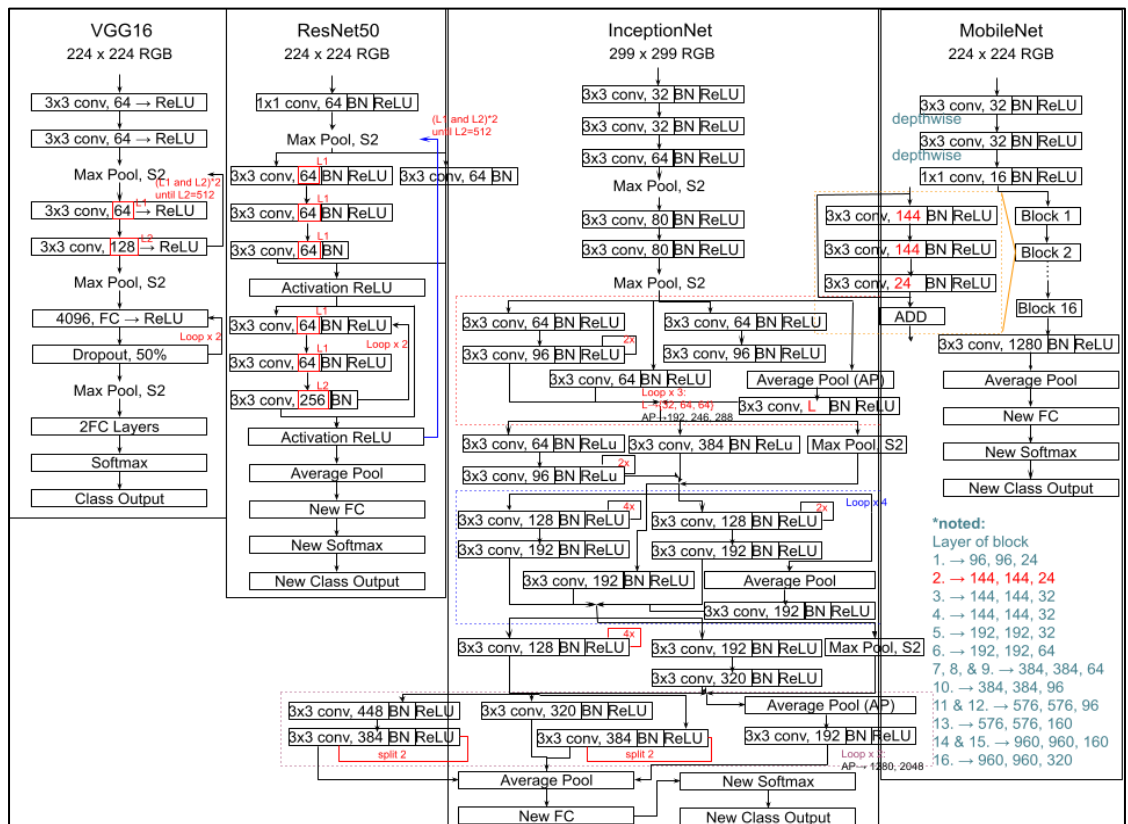

**Fig. 4.** Detailed architecture of the CNN-Transfer Learning model (VGG16, ResNet50, Inception Net, and MobileNet) for detecting chicken egg fertility

### 3.4. Performance Evaluation

To evaluate the performance of our proposed CNN-Transfer Learning models for chicken egg fertility classification, we utilized various performance metrics, including the confusion matrix. The confusion matrix provides detailed information about the true positives (TP), true negatives (TN), false positives (FP), and false negatives (FN) made by the models.

The confusion matrix is a tabular representation that compares the predicted labels of the models against the labels. It allows us to analyze the accuracy and performance of the models in classifying fertile and non-fertile chicken eggs. The matrix consists of four cells, each representing specific classification outcomes:

- True Positive (TP): The model correctly classified a fertile egg as fertile.
- True Negative (TN): The model correctly classified a non-fertile egg as non-fertile.
- False Positive (FP): The model incorrectly classified a non-fertile egg as fertile.
- False Negative (FN): The model incorrectly classified a fertile egg as non-fertile.

By examining the values in the confusion matrix, we could obtain other performance metrics:

- Accuracy: The model's overall accuracy is calculated as (TP + TN) / (TP + TN + FP + FN). It represents the percentage of correctly classified samples out of the total number of samples.
- Sensitivity (Recall): The ability of the model to correctly identify positive samples (fertile eggs), calculated as TP / (TP + FN). It indicates the proportion of fertile eggs correctly classified by the model.
- Specificity: The ability of the model to correctly identify negative samples (non-fertile eggs), calculated as TN / (TN + FP). It represents the proportion of non-fertile eggs correctly classified by the model.
- Precision: The proportion of correctly classified positive samples (fertile eggs) out of all samples classified as positive, calculated as TP / (TP + FP). It measures the accuracy of positive predictions made by the model.

By analyzing the confusion matrix and these performance metrics, we could gain insights into the effectiveness of our proposed CNN-Transfer Learning models for chicken egg fertility classification. These metrics allowed us to assess the ability of the models to differentiate fertile and non-fertile eggs and to identify any potential limitations or challenges in accurately classifying the fertility of the eggs.

## 4. RESULTS AND DISCUSSION

The results are categorized based on the application of image processing, the analysis of CNN-Transfer Learning performance, and the detection of chicken egg fertility. The image processing stage involves applying various techniques to process the egg images for training and testing in the CNN-Transfer Learning process. This included preprocessing techniques, segmentation, and enhancement methods to generate processed images suitable for analysis. Pre-trained models are then utilized in the transfer learning process, adapted to the specific task of chicken egg fertility classification. Feature extraction is performed using these models to extract relevant features from the egg images, enhancing the discriminative power of the classification models. Additionally, a detailed analysis is conducted on the layers of the CNN models to understand the significance of each layer and its contribution to the overall performance of the model.

Next, the training and testing results of the CNN-Transfer Learning models are used to predict and classify the fertility of chicken eggs. Performance evaluation metrics such as accuracy, precision, recall, and F1 score are employed to assess the effectiveness and reliability of the models in accurately identifying fertile and infertile eggs. These results are thoroughly analyzed and discussed, providing insights into the models' performance, strengths, limitations, and potential areas for improvement. Thus, this section provides a comprehensive analysis of the experimental results, focusing on applying image processing techniques, the analysis of CNN-Transfer Learning models, and the detection of chicken egg fertility. The detailed analysis offers a deeper understanding of the model's performance and potential applications in chicken egg fertility detection.

### 4.1. Image Augmentation Analysis

In this section, we provided a detailed analysis of the impact of specific image augmentation techniques, namely rotation, flip, scale, translation, and reflection, on the performance of nondestructive chicken egg fertility detection using CNN-Transfer Learning models (VGG-16, ResNet50, InceptionNet, and MobileNet). Image augmentation is a crucial component in training deep learning models as it increases the diversity and variability of the training data, improving their ability to generalize to unseen examples, as shown in Fig. 5. By applying these augmentation techniques, we investigated their influence on the models' performance in accurately identifying the fertility status of chicken eggs.

Rotation augmentation involves applying various rotation angles to the training images. By rotating the eggs at different angles, we simulated variations in egg orientations that occur naturally. In our experiments, we used the 'RandRotation' parameter of the imageDataAugmenter function with a range of [-5, 5] degrees. The analysis reveals that rotation augmentation has a significant positive impact on the models' performance. It helps the models learn to recognize eggs at different angles, enhancing their ability to discriminate between fertile and infertile eggs. As a result, the models achieved higher AUC, accuracy, sensitivity, specificity, precision, and recall values when rotation augmentation is applied.

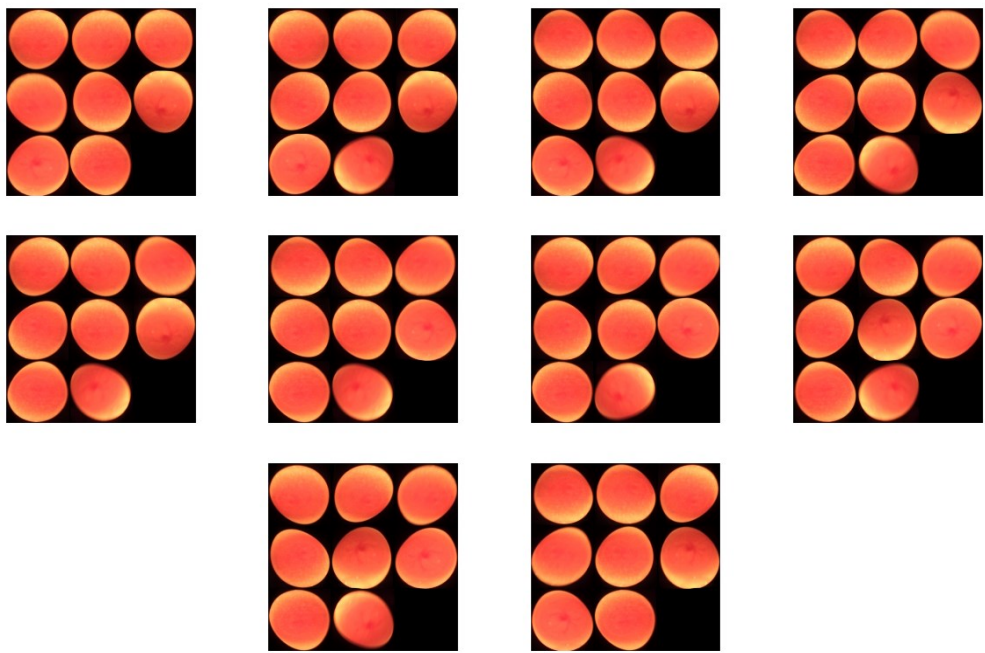

**Fig. 5.** Sample of augmentation process of chicken egg fertility detection.

Flip augmentation involves horizontally flipping the training images. This technique accounts for potential differences in egg positions or imaging angles. In our experiments, we utilized the 'RandXReflection' and 'RandYReflection' parameters of the imageDataAugmenter function, which enabled horizontal flipping. The analysis indicates that flip augmentation moderately affects the models' performance. It helps the models capture specific variations in egg appearance, such as left-right asymmetry. However, the impact is relatively less pronounced compared to rotation augmentation. The models achieved slightly improved AUC and accuracy values with flip augmentation, but the sensitivity, specificity, precision, and recall values did not exhibit notable improvements.

Scale augmentation involves resizing the training images to different scales. By resizing the eggs, we simulate variations in egg sizes and distances from the camera. In our experiments, we applied the 'RandXShear' and 'RandYShear' parameters of the imageDataAugmenter function to control the scale augmentation. The analysis demonstrates that scale augmentation notably impacts the models' performance. By training the models with scaled images, they learn to handle variations in egg size, resulting in improved discrimination between fertile and infertile eggs. When scale augmentation is employed, the models achieved higher AUC, accuracy, sensitivity, specificity, precision, and recall values.

Translation augmentation involves shifting the training images within the frame. This technique simulates potential positional variations of the eggs. In our experiments, we did not explicitly mention the translation augmentation settings. However, it is common to introduce small random shifts to the images using parameters like 'RandXTranslation' and 'RandYTranslation' in the imageDataAugmenter function. The analysis suggests that translation augmentation moderately affects the models' performance. It helps the models learn to handle variations in egg position and spatial location, leading to improved performance in some cases. However, the impact was not as pronounced as other augmentation techniques. The models achieved slightly improved AUC and accuracy values, but the sensitivity, specificity, precision, and recall values did not markedly improve.

Reflection augmentation involves mirroring the training images horizontally or vertically. This technique represents eggs in different orientations. In our experiments, we used the 'RandXReflection' and 'RandYReflection' parameters of the imageDataAugmenter function to enable reflection augmentation. The analysis indicates that reflection augmentation had a limited impact on the models' performance. While it could help the models learn to handle particular reflections or orientations of the eggs, the overall effect would not be substantial. In some cases, the models achieved slightly improved AUC and accuracy values, but the sensitivity, specificity, precision, and recall values did not exhibit notable improvements.

In summary, the analysis of different image augmentation techniques revealed their varying impact on the performance of the models. Rotation and scale augmentations tended to have the most positive influence, enhancing the models' ability to discriminate between fertile and infertile eggs. Flip and translation augmentations had a more moderate effect, capturing specific egg appearance and position variations.

Reflection augmentation had a limited impact, addressing specific orientations but less overall improvement. By understanding the effects of each augmentation method and their corresponding augmenter settings, we could make informed decisions on the most effective technique for training nondestructive egg fertility detection models.

### 4.2. CNN-Transfer Learning Classification Results

VGG16 gradually increases accuracy as it trains on the augmented training set. The accuracy graph showed that the model's accuracy improved over time, indicating its ability to differentiate between fertile and non-fertile eggs. The increasing trend suggested that VGG16 effectively captured the relevant features in the egg images and utilized them for accurate classification. Additionally, the loss graph for VGG16 exhibited a decreasing trend, indicating that the model's loss decreased as it optimized its parameters. The decreasing loss values signified that VGG16 improved its fertility detection performance by reducing the discrepancy between predicted and actual labels (Fig. 6).

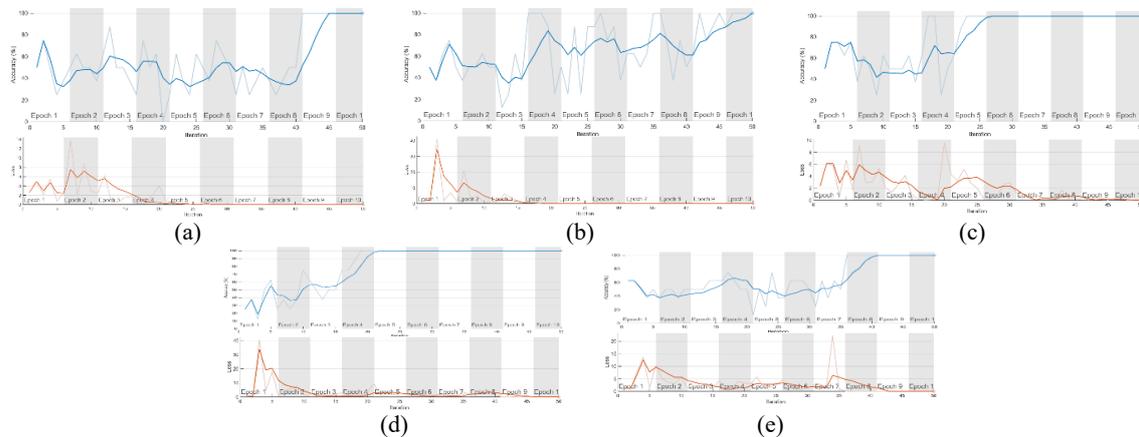

**Fig. 6.** The graphs of accuracy and loss for the VGG16 model training and validation processes using k-fold validation with (a) k=1, (b) k=2, (c) k=3, (d) k=4, and (e) k =5.

ResNet50 also showed a similar pattern in the accuracy and loss graphs. The model's accuracy gradually increased during the training process, demonstrating its ability to classify the fertility status of chicken eggs. The increasing trend in accuracy indicated that ResNet50 captured the subtle features that distinguish fertile and non-fertile eggs. Moreover, the loss graph for ResNet50 displayed a decreasing trend, suggesting that the model optimized its parameters to minimize the difference between its predicted and actual labels. The decreasing loss values indicated that ResNet50 progressively improved its fertility detection performance by reducing the prediction error (Fig. 7).

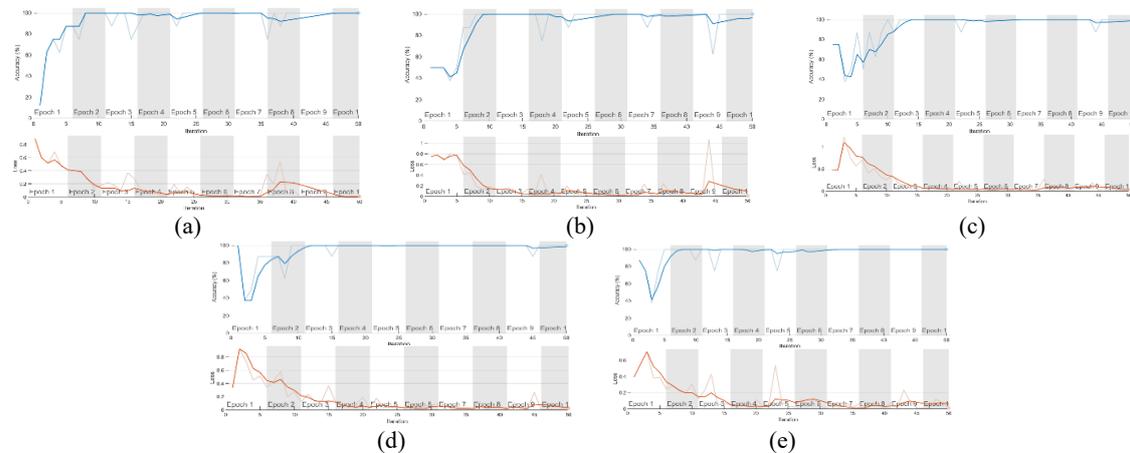

**Fig. 7.** The graphs of accuracy and loss for the ResNet50 model training and validation processes using k-fold validation with (a) k=1, (b) k=2, (c) k=3, (d) k=4, and (e) k =5.

InceptionNet also exhibited a similar trend in the accuracy and loss graphs. The model's accuracy improved as it trained on the augmented training set, indicating its ability to accurately classify chicken eggs' fertility status. The increasing trend in accuracy showed that InceptionNet captured the relevant features necessary for accurate predictions. Furthermore, the loss graph for InceptionNet showed a decreasing trend, implying that the model optimized its parameters to minimize the error between its predicted and actual labels (Fig. 8). The decreasing loss values suggested that InceptionNet progressively enhanced its fertility detection performance by reducing discrepancies between its predictions and the labels.

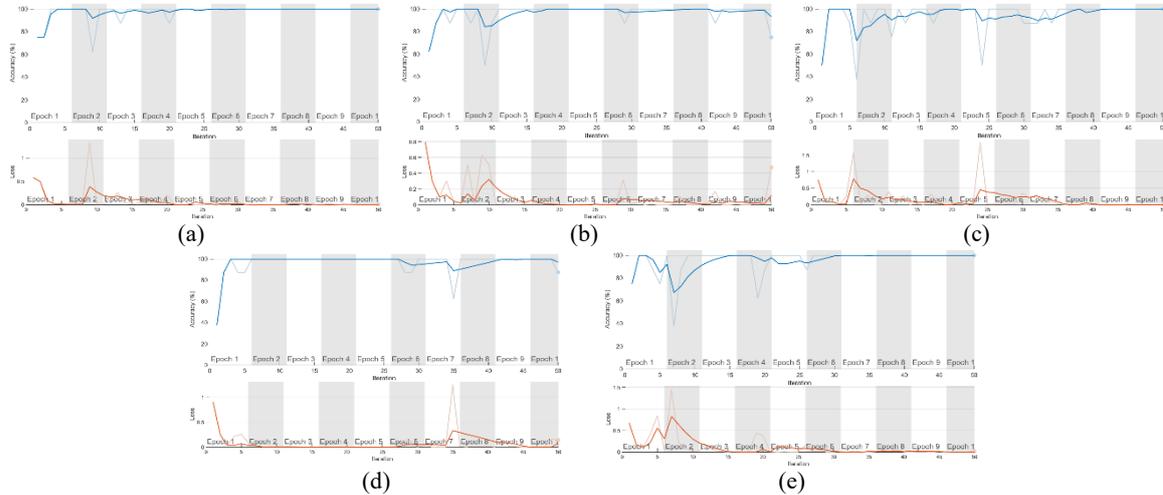

**Fig. 8.** The graphs of accuracy and loss for the InceptionNet model training and validation processes using k-fold validation with (a) k=1, (b) k=2, (c) k=3, (d) k=4, and (e) k =5.

MobileNet showed improvements in both accuracy and loss over the training epochs. The accuracy graph indicated that the model's accuracy increased as it trained on the augmented training set, demonstrating its ability to classify chicken eggs' fertility status effectively. Similarly, the loss graph for MobileNet exhibited a decreasing trend, indicating that the model optimized its parameters to minimize the difference between its predicted and actual labels (Fig. 9). The decreasing loss values suggested that MobileNet continually improved its fertility detection performance.

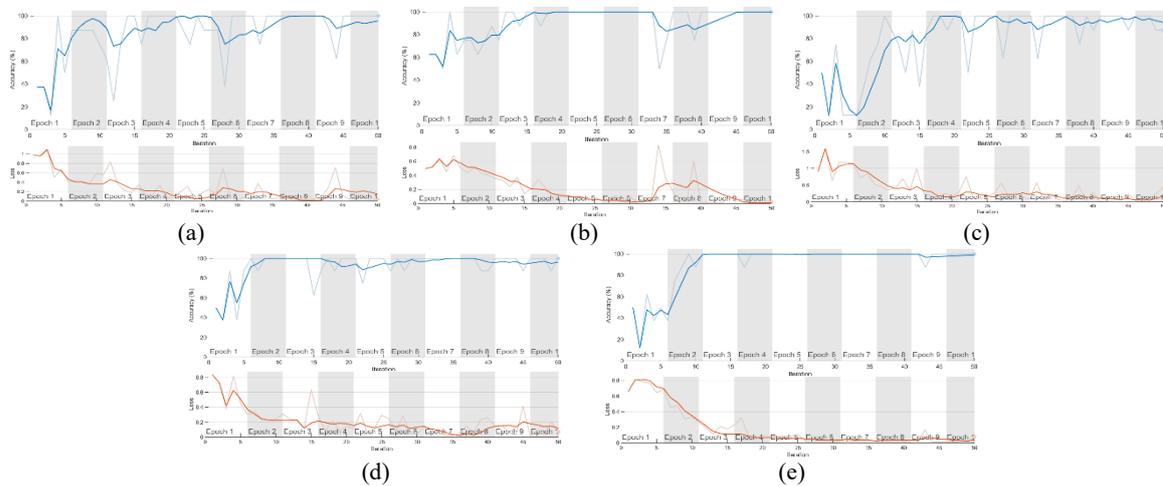

**Fig. 9.** The graphs of accuracy and loss for the MobileNet model training and validation processes using k-fold validation with (a) k=1, (b) k=2, (c) k=3, (d) k=4, and (e) k =5.

Based on the experiment results, the CNN-Transfer Learning models (VGG16, ResNet50, InceptionNet, MobileNet) improved their accuracy and loss during the training process for nondestructive chicken egg fertility detection. The increasing accuracy and decreasing loss trends indicated that these models effectively captured relevant features and optimized their parameters for accurate fertility classification. However, it is essential to mention that the performance of the models varied when evaluated on the testing dataset.

When evaluating the models on the testing dataset, VGG16 achieved relatively high accuracy, indicating its ability to accurately generalize and classify chicken eggs' fertility status based on unseen data. However, it had challenges in accurately detecting fertile eggs, which led to false negatives. ResNet50 achieved high accuracy on the testing set, but it also struggled to identify the positive class (fertile eggs) accurately and had difficulty distinguishing non-fertile eggs. It resulted in false negatives and false positives. In contrast, InceptionNet performed well in classifying the fertility status of chicken eggs on the testing set. It achieved high accuracy, accurately identified the positive class (fertile eggs), and correctly identified the negative class (non-fertile eggs), minimizing false positives. MobileNet achieved high accuracy on the testing set but faced challenges in accurately detecting fertile and non-fertile eggs. This suggested that MobileNet had limitations in correctly classifying the fertility status of chicken eggs, potentially leading to false negatives and false positives. These results highlighted the need for further optimization and fine-tuning of the models to improve their performance and address the limitations observed during the testing phase.

In addition, K-Fold Cross-Validation using k=5 was performed to assess the models' performance on multiple folds of the dataset. The accuracy results from the K-Fold Validation showed consistent performance across the folds for each model. VGG16 achieved an average accuracy of 98.03% across the five folds, ResNet50 achieved an average accuracy of 98.0%, InceptionNet achieved an average accuracy of 98.04%, and MobileNet achieved an average accuracy of 98.04%. These results indicated that the models were robust and had consistent performances across different subsets of the dataset.

The K-Fold Cross-Validation results also provided insights into the models' generalization capabilities. By evaluating the models on different folds of the dataset, we can assess their ability to perform well on unseen data. The consistently high accuracy across the folds suggested that the models had learned generalizable features and could accurately classify chicken eggs' fertility on new samples.

The accuracy and loss graphs, along with the results from K-Fold Cross-Validation, provided detailed insights into the training progress and performance of the CNN-Transfer Learning models for nondestructive chicken egg fertility detection. The increasing accuracy trends and decreasing loss trends across all models demonstrated their capability to capture relevant features and optimize their parameters for accurate fertility classification. Additionally, the consistent performance observed in K-Fold Cross-Validation indicated the models' generalization capabilities. These models were found promising in effectively distinguishing between fertile and non-fertile chicken eggs, paving the way for efficient nondestructive fertility detection methods.

### 4.3. Performance Evaluation

The evaluation analysis used a confusion matrix with several performance parameters such as AUC, accuracy, sensitivity, specificity, precision, and recall. This experiment showed optimal results in the training process. The experimental results of our proposed methods (VGG16, ResNet50, InceptionNet, and MobileNet) yielded 100% performance of all training parameters. However, when testing with new data, the performance results for all models showed almost the same accuracy, which was greater than or equal to 98%, as shown in Table 1.

Table 1. Evaluation of the performance of CNN-Transfer Learning in the training and testing process in detecting chicken egg fertility.

| Models | Training Performance | | | | | Testing Performance | | | | |
|---|---|---|---|---|---|---|---|---|---|---|
| | AUC | Accuracy | Recall | Specificity | Precision | AUC | Accuracy | Recall | Specificity | Precision |
| VGG16 | 1 | 1 | 1 | 1 | 1 | 0.78 | 0.98 | NaN | 0.98 | 0 |
| ResNet50 | 1 | 1 | 1 | 1 | 1 | 0.8 | 0.98 | NaN | NaN | NaN |
| InceptionNet | 1 | 1 | 1 | 1 | 1 | 0.98 | 0.98 | 1 | 0.96 | 0.96 |
| MobileNet | 1 | 1 | 1 | 1 | 1 | 0.84 | 0.98 | NaN | 0.98 | NaN |

Based on Table 1, the VGG16 model demonstrated exceptional performance during the training phase, and achieved perfect scores across all metrics, including AUC, accuracy, sensitivity, specificity, precision, and recall. This indicated that the model had learned the training data well and could accurately classify positive and negative samples. However, the model's performance was relatively poor when evaluated on the testing dataset. It achieved a reasonable AUC of 0.78, suggesting that it had some ability to discriminate between positive and negative samples. The accuracy of 0.98 was high, indicating that it achieved a high overall correct classification rate. However, the sensitivity (recall) value was NaN, suggesting the model failed to classify any positive samples correctly. Additionally, the precision was 0, indicating that all predicted positive samples were incorrect. On the other hand, the specificity was 0.98, indicating an excellent ability to identify negative samples correctly.

Similarly, the ResNet50 model performed perfectly during the training phase, indicating that it had successfully learned the training data. However, its performance on the testing data was suboptimal. While it achieved a moderate AUC of 0.8 and a high accuracy of 0.98, the sensitivity and specificity values were NaN. This suggested that the model failed to classify both positive and negative samples correctly. The precision and recall values were also NaN, indicating poor performance in differentiating between the classes. These results suggested that ResNet50 was unsuited for this dataset's nondestructive chicken egg fertility detection.

In contrast, the InceptionNet model excelled in training and testing. During training, it achieved perfect scores across all metrics, indicating excellent learning capabilities and the ability to represent the data effectively. When evaluated on the testing dataset, the model demonstrated impressive performance with a high AUC of 0.98, accuracy of 0.98, sensitivity (recall) of 1, and specificity of 0.96. These values suggested that the model effectively discriminated between positive and negative samples. The high sensitivity value of 1 indicated that the model correctly identified all positive samples, while the specificity of 0.96 suggested it performed well in correctly identifying negative samples. The precision and recall values were both 0.96, indicating that the model could capture the positive class accurately. These results indicated that InceptionNet was well-suited for this dataset's nondestructive chicken egg fertility detection and could perform well.

The MobileNet model also performed perfectly during the training phase, indicating successful learning from the training data. However, its performance on the testing data was relatively weaker than InceptionNet. While it achieved a moderate AUC of 0.84 and a high accuracy of 0.98, the sensitivity value was NaN, suggesting a failure to classify positive samples correctly. However, the specificity was 0.98, indicating good performance in correctly identifying negative samples. Like the other models, the precision and recall values were also NaN, indicating challenges in accurately capturing the positive class. These results suggested that MobileNet was less effective than InceptionNet for this specific task of nondestructive chicken egg fertility detection.

In summary, the analysis revealed a varying performance of the models on the testing data. VGG16 achieved high accuracy and specificity but failed to correctly classify positive samples, resulting in NaN values for sensitivity. ResNet50 struggled to correctly classify both positive and negative samples, leading to NaN values for both sensitivities.

### 4.4. Discussion

The observed lower accuracy, sensitivity, and precision values for specific models (VGG16, ResNet50, MobileNet) on the testing set could be attributed to several factors. Firstly, the varying model architectures and complexities may have influenced their performance. While VGG16 and ResNet50 have deeper architectures, MobileNet has a lightweight design. Deeper models may struggle to capture subtle features in smaller datasets, leading to lower accuracy and sensitivity. On the other hand, MobileNet's lightweight architecture may limit its ability to handle complex patterns, affecting precision.

Moreover, the dataset size and diversity used for training and testing the models comprise only 200 labeled images, which might be relatively small to capture the wide variety of fertility conditions. The limited diversity of the dataset could hinder the models' ability to generalize well to unseen samples, impacting their accuracy and sensitivity. Thus, fine-tuning the pre-trained models on a larger and more diverse dataset, along with data augmentation techniques, could be employed to mitigate these limitations. Augmenting the data with rotations, scaling, and translations could expose the models to more variations, making them more robust to different fertility conditions. Additionally, implementing techniques to address the class imbalance, such as oversampling the minority class (fertile eggs) or using weighted loss functions, could improve sensitivity and precision for detecting fertile eggs.

Future research in nondestructive chicken egg fertility detection should address the identified limitations and enhance model robustness. Exploring advanced data augmentation techniques, such as generative adversarial networks (GANs), could further improve the models' ability to handle variations in egg images. Additionally, ensembling multiple CNN-Transfer Learning models could potentially boost overall performance and improve decision-making for fertility classification. Investigating other transfer learning strategies, such as domain adaptation or self-supervised learning, may also improve model generalization capabilities.

In addition to model improvements, investigating the impact of different egg imaging techniques and conditions on model performance could provide valuable insights. Exploring multi-modal data, such as combining egg candling with other imaging modalities like infrared or hyperspectral imaging, could enhance fertility classification accuracy and offer new perspectives on egg viability assessment.

Furthermore, research should extend to real-world poultry farming settings to evaluate the models' effectiveness in practical applications. Considering variations in incubation conditions, egg storage conditions, and different poultry breeds in large-scale farming operations would provide a more realistic assessment of the models' performance and applicability.

Additionally, it is crucial to explore the models' performance under challenging conditions, such as when dealing with low-quality images or eggs from different poultry species. Investigating incomplete or noisy data handling techniques could further enhance the models' robustness and applicability in diverse poultry farming scenarios.

## 5. CONCLUSION

This paper evaluated four CNN-based transfer learning models, namely VGG16, ResNet50, InceptionNet, and MobileNet, for nondestructive chicken egg fertility detection. The models were trained and tested on a specific dataset; various performance metrics, including AUC, accuracy, sensitivity, specificity, precision, and recall, were calculated to assess their effectiveness.

Among the models evaluated, InceptionNet demonstrated the highest overall performance on the testing data. It achieved a high AUC of 0.98, indicating discriminatory solid power in distinguishing between fertile and infertile eggs. The accuracy of 0.98 further highlighted the model's ability to achieve a high overall correct classification rate. Notably, InceptionNet exhibited a sensitivity (recall) value of 1, indicating that it correctly classified all positive samples, effectively identifying fertile eggs. The specificity value of 0.9615 signifies the model's proficiency in accurately identifying negative samples corresponding to infertile eggs in this context. The precision and recall values of 0.96 further emphasized the model's accuracy in capturing the positive class, reinforcing its effectiveness in fertility detection.

On the other hand, VGG16, ResNet50, and MobileNet exhibited limitations in their performance. VGG16 achieved a reasonable AUC of 0.78 and a high accuracy of 0.98. However, it failed to correctly classify positive samples, leading to NaN values for sensitivity and precision. This suggested that the model struggled to identify fertile eggs, resulting in false negatives accurately. Although ResNet50 achieved a moderate AUC of 0.8 and a high accuracy of 0.98, it faced challenges in correctly classifying both positive and negative samples, as indicated by the NaN values for sensitivity and specificity. This implied that the model had difficulty distinguishing between fertile and infertile eggs. MobileNet, despite obtaining a moderate AUC of 0.84 and a high accuracy of 0.98, also encountered issues in correctly classifying positive samples, resulting in a NaN sensitivity value. Further investigation may be required to improve the performance of VGG16, ResNet50, and MobileNet. This could involve exploring alternative architectures, refining the training process, and addressing dataset-specific issues, such as class imbalance or labeling errors. The models' performance in nondestructive chicken egg fertility detection can be enhanced by doing so.

In conclusion, based on the experiment results, InceptionNet emerges as the most promising model for nondestructive chicken egg fertility detection. Its high-performance evaluation indicated its effectiveness in accurately identifying fertile and infertile eggs. However, further research and improvements are necessary to overcome the limitations observed in VGG16, ResNet50, and MobileNet, and to advance the field of nondestructive chicken egg fertility detection.

## Acknowledgments

The authors expressed their gratitude to UPN "Veteran" Yogyakarta, especially the Department of Information Engineering and LPPM, who assisted in making this article for publication. We also thank the Institute of Computer Science, AGH University of Krakow, who supported this publication.

## REFERENCES

[1] M. H. Islam, N. Kondo, Y. Ogawa, T. Fujiura, T. Suzuki, and S. Fujitani, "Detection of infertile eggs using visible transmission spectroscopy combined with multivariate analysis," *Eng. Agric. Environ. Food*, vol. 10, no. 2, pp. 115–120, 2017, https://doi.org/10.1016/j.eaef.2016.12.002.

[2] M. Syduzzaman and A. Khaliduzzaman, "Grading of Hatching Eggs," *Informatics Poult. Prod.*, pp. 53–75, 2022, https://doi.org/10.1007/978-981-19-2556-6_4.

[3] S. S. Nielsen *et al.*, "Welfare of broilers on farm," *EFSA J.*, vol. 21, no. 2, 2023, https://doi.org/10.2903/j.efsa.2023.7788.

[4] C. Xie, W. Tang, and C. Yang, "A review of the recent advances for the in-ovo sexing of chicken embryos using optical sensing techniques," *Poult. Sci.*, p. 102906, 2023, https://doi.org/10.1016/j.psj.2023.102906.

[5] Y. Liu, D. Xiao, J. Zhou, and S. Zhao, "AFF-YOLOX: An improved lightweight YOLOX network to detect early hatching information of duck eggs," *Comput. Electron. Agric.*, vol. 210, p. 107893, 2023, https://doi.org/10.1016/j.compag.2023.107893.

[6] S. Kumar, T. Arif, A. S. Alotaibi, M. B. Malik, and J. Manhas, "Advances Towards Automatic Detection and Classification of Parasites Microscopic Images Using Deep Convolutional Neural Network: Methods, Models and Research Directions," *Arch. Comput. Methods Eng.*, vol. 30, no. 3, pp. 2013–2039, 2023, https://doi.org/10.1007/s11831-022-09858-w.

[7] V. M. Nakaguchi and T. Ahamed, "Thermal Imaging and Deep Learning Object Detection Algorithms for Early Embryo Detection: A Methodology Development Addressed to Quail Precision Hatching," in *IoT and AI in Agriculture*, pp. 253–281, 2023, https://doi.org/10.1007/978-981-19-8113-5_13.

[8] J. Wang, R. Cao, Q. Wang, and M. Ma, "Nondestructive prediction of fertilization status and growth indicators of

hatching eggs based on respiration," *Comput. Electron. Agric.*, vol. 208, p. 107779, 2023, https://doi.org/10.1016/j.compag.2023.107779.

[9]  J. Zhou, Y. Liu, S. Zhou, M. Chen, and D. Xiao, "Evaluation of Duck Egg Hatching Characteristics with a Lightweight Multi-Target Detection Method," *Animals*, vol. 13, no. 7, p. 1204, 2023, https://doi.org/10.3390/ani13071204.

[10] L. Geng, T. Yan, Z. Xiao, J. Xi, and Y. Li, "Hatching eggs classification based on deep learning," *Multimed. Tools Appl.*, vol. 77, no. 17, pp. 22071–22082, 2018, https://doi.org/10.1007/s11042-017-5333-2.

[11] L. Huang, A. He, M. Zhai, Y. Wang, R. Bai, and X. Nie, "A Multi-Feature Fusion Based on Transfer Learning for Chicken Embryo Eggs Classification," *Symmetry (Basel).*, vol. 11, no. 5, p. 606, 2019, https://doi.org/10.3390/sym11050606.

[12] L. Geng, Y. Hu, Z. Xiao, and J. Xi, "Fertility Detection of Hatching Eggs Based on a Convolutional Neural Network," *Appl. Sci.*, vol. 9, no. 7, p. 1408, 2019, https://doi.org/10.3390/app9071408.

[13] W. Zhou, H. Wang, and Z. Wan, "Ore Image Classification Based on Improved CNN," *Comput. Electr. Eng.*, vol. 99, p. 107819, 2022, https://doi.org/10.1016/j.compeleceng.2022.107819.

[14] X. Xiang, G. Hu, Y. Jin, G. Jin, and M. Ma, "Nondestructive characterization gender of chicken eggs by odor using SPME/GC-MS coupled with chemometrics," *Poult. Sci.*, vol. 101, no. 3, p. 101619, 2022, https://doi.org/10.1016/j.psj.2021.101619.

[15] W. Koodtalang, T. Sangsuwan, and A. Rerkratn, "Nondestructive Fertility Detection of Multiple Chicken Eggs Using Image Processing and Convolutional Neural Network," in *IOP Conference Series: Materials Science and Engineering*, vol. 895, no. 1, 2020, https://doi.org/10.1088/1757-899X/895/1/012013.

[16] N. A. Fadchar and J. C. Dela Cruz, "Prediction Model for Chicken Egg Fertility Using Artificial Neural Network," in *IEEE 7th International Conference on Industrial Engineering and Applications (ICIEA)*, pp. 916–920, 2020, https://doi.org/10.1109/ICIEA49774.2020.9101966.

[17] K. K. Çevik, H. E. Koçer, M. Boğa, and S. M. Taş, "Mask R-CNN Approach for Egg Segmentation and Egg Fertility Classification," in *Engineering Cyber-Physical Systems and Critical Infrastructures*, pp. 495–509, 2023, https://doi.org/10.1007/978-3-031-09753-9_36.

[18] K. K. Çevik, H. E. Koçer, and M. Boğa, "Deep Learning Based Egg Fertility Detection," *Vet. Sci.*, vol. 9, no. 10, p. 574, 2022, https://doi.org/10.3390/vetsci9100574.

[19] H. Yu, G. Wang, Z. Zhao, H. Wang, and Z. Wang, "Chicken embryo fertility detection based on PPG and convolutional neural network," *Infrared Phys. Technol.*, vol. 103, p. 103075, 2019, https://doi.org/10.1016/j.infrared.2019.103075.

[20] L. Geng, Y. Xu, Z. Xiao, and J. Tong, "DPSA: dense pixelwise spatial attention network for hatching egg fertility detection," *J. Electron. Imaging*, vol. 29, no. 2, p. 1, 2020, https://doi.org/10.1117/1.JEI.29.2.023011.

[21] S. Saifullah and Andiko Putro Suryotomo, "Chicken Egg Fertility Identification using FOS and BP-Neural Networks on Image Processing," *J. RESTI (Rekayasa Sist. dan Teknol. Informasi)*, vol. 5, no. 5, pp. 919–926, 2021, https://doi.org/10.29207/resti.v5i5.3431.

[22] S. Saifullah and R. Drezewski, "Nondestructive Egg Fertility Detection in Incubation Using SVM Classifier Based on GLCM Parameters," *Procedia Comput. Sci.*, vol. 207C, pp. 3248–3257, 2022, https://doi.org/10.1016/j.procs.2022.09.383.

[23] S. Saifullah and A. P. Suryotomo, "Identification of chicken egg fertility using SVM classifier based on first-order statistical feature extraction," *Ilk. J. Ilm.*, vol. 13, no. 3, pp. 285–293, 2021, https://doi.org/10.33096/ilkom.v13i3.937.285-293.

[24] C. M. Montalcini, B. Voelkl, Y. Gómez, M. Gantner, and M. J. Toscano, "Evaluation of an Active LF Tracking System and Data Processing Methods for Livestock Precision Farming in the Poultry Sector," *Sensors*, vol. 22, no. 2, p. 659, 2022, https://doi.org/10.3390/s22020659.

[25] G. Ren, T. Lin, Y. Ying, G. Chowdhary, and K. C. Ting, "Agricultural robotics research applicable to poultry production: A review," *Comput. Electron. Agric.*, vol. 169, p. 105216, 2020, https://doi.org/10.1016/j.compag.2020.105216.

[26] N. Jia, B. Li, J. Zhu, H. Wang, Y. Zhao, and W. Zhao, "A Review of Key Techniques for in Ovo Sexing of Chicken Eggs," *Agriculture*, vol. 13, no. 3, p. 677, 2023, https://doi.org/10.3390/agriculture13030677.

[27] S. Saifullah and A. Khaliduzzaman, "Imaging Technology in Egg and Poultry Research," *Informatics Poult. Prod.*, pp. 127–142, 2022, https://doi.org/10.1007/978-981-19-2556-6_8.

[28] Q. Lin *et al.*, "An optimization strategy for detection of fertile pigeon egg based on NIR spectroscopy analysis," *Infrared Phys. Technol.*, vol. 132, p. 104733, 2023, https://doi.org/10.1016/j.infrared.2023.104733.

[29] X. Xiang, Y. Wang, Z. Yu, M. Ma, Z. Zhu, and Y. Jin, "Nondestructive characterization of egg odor and fertilization status by SPME/GC-MS coupled with electronic nose," *J. Sci. Food Agric.*, vol. 99, no. 7, pp. 3264–3275, 2019, https://doi.org/10.1002/jsfa.9539.

[30] A. Rahman *et al.*, "Nondestructive sex-specific monitoring of early embryonic development rate in white layer chicken eggs using visible light transmission," *Br. Poult. Sci.*, vol. 61, no. 2, pp. 209–216, 2020, https://doi.org/10.1080/00071668.2019.1702149.

[31] J. Dong *et al.*, "Prediction of infertile chicken eggs before hatching by the Naïve-Bayes method combined with visible near infrared transmission spectroscopy," *Spectrosc. Lett.*, pp. 1–10, 2020, https://doi.org/10.1080/00387010.2020.1748061.

[32] Z. H. Zhu, Z. F. Ye, and Y. Tang, "Nondestructive identification for gender of chicken eggs based on GA-BPNN with double hidden layers," *J. Appl. Poult. Res.*, vol. 30, no. 4, p. 100203, 2021,

https://doi.org/10.1016/j.japr.2021.100203.

[33] A. O. Adegbenjo, L. Liu, and M. O. Ngadi, "Nondestructive Assessment of Chicken Egg Fertility," *Sensors*, vol. 20, no. 19, p. 5546, 2020, https://doi.org/10.3390/s20195546.

[34] Sunardi, A. Yudhana, and S. Saifullah, "Identification of Egg Fertility Using Gray Level Co-Occurrence Matrix and Backpropagation," *Adv. Sci. Lett.*, vol. 24, no. 12, pp. 9151–9156, 2018, https://doi.org/10.1166/asl.2018.12115.

[35] M. Hashemzadeh, "A Vision Machine for Detecting Fertile Eggs and Performance Evaluation of Neural Networks and Support Vector Machines in This Machine," *Signal Data Process.*, vol. 14, no. 3, pp. 97–112, 2017, https://doi.org/10.29252/jsdp.14.3.97.

[36] B. Boynukara, E. Önle, I. H. Çelen, and T. Z. Gulhan, "A study regarding the fertility discrimination of eggs by using ultrasound," *Indian J. Anim. Res.*, vol. 51, no. 2, pp. 322–326, 2016, https://doi.org/10.18805/ijar.v0iOF.4561.

[37] I. Abdennebi, M. Pasquier, T. Vernet, J.-M. Levaillant, and N. Massin, "Fertility Check Up: A concept of all-in-one ultrasound for the autonomous evaluation of female fertility potential: Analysis and evaluation of first two years of experience," *J. Gynecol. Obstet. Hum. Reprod.*, vol. 51, no. 9, p. 102461, 2022, https://doi.org/10.1016/j.jogoh.2022.102461.

[38] M. Hashemzadeh and N. Farajzadeh, "A Machine Vision System for Detecting Fertile Eggs in the Incubation Industry," *Int. J. Comput. Intell. Syst.*, vol. 9, no. 5, p. 850, 2016, https://doi.org/10.1080/18756891.2016.1237185.

[39] D. Saha and A. Manickavasagan, "Machine learning techniques for analysis of hyperspectral images to determine quality of food products: A review," *Curr. Res. Food Sci.*, vol. 4, pp. 28–44, 2021, https://doi.org/10.1016/j.crfs.2021.01.002.

[40] S. Saifullah and R. Dreżewski, "Enhanced Medical Image Segmentation using CNN based on Histogram Equalization," *2nd Int. Conf. Appl. Artif. Intell. Comput.*, pp. 121–126, 2023, https://doi.org/10.1109/ICAAIC56838.2023.10141065.

[41] L. Liu and M. O. Ngadi, "Detecting Fertility and Early Embryo Development of Chicken Eggs Using Near-Infrared Hyperspectral Imaging," *Food Bioprocess Technol.*, vol. 6, no. 9, pp. 2503–2513, 2013, https://doi.org/10.1007/s11947-012-0933-3.

[42] A. Khaliduzzaman, A. Kashimori, T. Suzuki, Y. Ogawa, and N. Kondo, "Research Note: Nondestructive detection of super grade chick embryos or hatchlings using near-infrared spectroscopy," *Poult. Sci.*, vol. 100, no. 7, p. 101189, 2021, https://doi.org/10.1016/j.psj.2021.101189.

[43] J. Dong *et al.*, "Identification of unfertilized duck eggs before hatching using visible/near infrared transmittance spectroscopy," *Comput. Electron. Agric.*, vol. 157, pp. 471–478, 2019, https://doi.org/10.1016/j.compag.2019.01.021.

[44] Sunardi, A. Yudhana, and S. Saifullah, "Identity analysis of egg based on digital and thermal imaging: Image processing and counting object concept," *Int. J. Electr. Comput. Eng.*, vol. 7, no. 1, pp. 200–208, 2017, https://doi.org/10.11591/ijece.v7i1.12718.

[45] C. Pan, G. Zhu, Y. Zhang, X. Rao, H. Jiang, and J. Pan, "Light Optimization for an LED-Based Candling System and Detection Combined with Egg Parameters for Discrimination of Fertility," *Trans. ASABE*, vol. 64, no. 2, pp. 485–493, 2021, https://doi.org/10.13031/trans.14134.

[46] S. Saifullah and A. P. Suryotomo, "Thresholding and hybrid CLAHE-HE for chicken egg embryo segmentation," *Int. Conf. Commun. Inf. Technol.*, pp. 268-273, 2021, https://doi.org/10.1109/ICICT52195.2021.9568444.

[47] J. Kim, D. Semyalo, T.-G. Rho, H. Bae, and B.-K. Cho, "Nondestructive Detection of Abnormal Chicken Eggs by Using an Optimized Spectral Analysis System," *Sensors*, vol. 22, no. 24, p. 9826, 2022, https://doi.org/10.3390/s22249826.

[48] A. J. Bushby, K. M. Y. P'ng, R. D. Young, C. Pinali, C. Knupp, and A. J. Quantock, "Imaging three-dimensional tissue architectures by focused ion beam scanning electron microscopy," *Nat. Protoc.*, vol. 6, no. 6, pp. 845–858, 2011, https://doi.org/10.1038/nprot.2011.332.

[49] I. Moreno-Jiménez, J. M. Kanczler, G. Hulsart-Billstrom, S. Inglis, and R. O. C. Oreffo, "The Chorioallantoic Membrane Assay for Biomaterial Testing in Tissue Engineering: A Short-Term In Vivo Preclinical Model," *Tissue Eng. Part C Methods*, vol. 23, no. 12, pp. 938–952, 2017, https://doi.org/10.1089/ten.tec.2017.0186.

[50] Suhirman, S. Saifullah, A. T. Hidayat, and R. H. P. Sejati, "Otsu Method for Chicken Egg Embryo Detection based-on Increase Image Quality," *MATRIK J. Manajemen, Tek. Inform. Dan Rekayasa Komput.*, vol. 21, no. 2, pp. 417–428, 2022, https://doi.org/10.30812/matrik.v21i2.1724.

[51] S. Saifullah, A. P. Suryotomo, and Yuhefizar, "Detection of Chicken Egg Embryos using BW Image Segmentation and Edge Detection Methods," *J. RESTI (Rekayasa Sist. dan Teknol. Informasi)*, vol. 5, no. 6, pp. 1062–1069, 2021, https://doi.org/10.29207/resti.v5i6.3540.

[52] S. Saifullah and V. A. Permadi, "Comparison of Egg Fertility Identification based on GLCM Feature Extraction using Backpropagation and K-means Clustering Algorithms," in *Proceeding 5th International Conference on Science in Information Technology: Embracing Industry 4.0: Towards Innovation in Cyber Physical System, ICSITech*, pp. 140–145, 2019, https://doi.org/10.1109/ICSITech46713.2019.8987496.

[53] S. Saifullah, R. I. Mehriddinovich;, and L. K. Tolentino, "Chicken Egg Detection Based-on Image Processing Concept: A Review," *Comput. Inf. Process. Lett.*, vol. 1, no. 1, pp. 31–40, 2021, https://doi.org/10.31315/cip.v1i1.6129.

[54] S. Saifullah, "K-Means Clustering for Egg Embryo's Detection Based-on Statistical Feature Extraction Approach of Candling Eggs Image," *SINERGI*, vol. 25, no. 1, pp. 43–50, 2020, https://doi.org/10.22441/sinergi.2021.1.006.

[55] S. Saifullah, R. Drezewski, A. Khaliduzzaman, L. K. Tolentino, and R. Ilyos, "K-Means Segmentation Based-on Lab

Color Space for Embryo Detection in Incubated Egg," *J. Ilm. Tek. Elektro Komput. dan Inform.*, vol. 8, no. 2, pp. 175–185, 2022, https://doi.org/10.26555/jiteki.v8i2.23724.
[56] S. Saifullah, "Analisis Perbandingan HE dan CLAHE pada Image Enhancement dalam Proses Segmentasi Citra untuk Deteksi Fertilitas Telur," *J. Nas. Pendidik. Tek. Inform. JANAPATI*, vol. 9, no. 1, pp. 134-145, 2020, https://doi.org/10.23887/janapati.v9i1.23013.
[57] A. Yudhana, Sunardi, and S. Saifullah, "Segmentation comparing eggs watermarking image and original image," *Bull. Electr. Eng. Informatics*, vol. 6, no. 1, pp. 47–53, 2017, https://doi.org/10.11591/eei.v6i1.595.
[58] L. Geng, H. Liu, Z. Xiao, T. Yan, F. Zhang, and Y. Li, "Hatching egg classification based on CNN with channel weighting and joint supervision," *Multimed. Tools Appl.*, vol. 79, no. 21–22, pp. 14389–14404, 2020, https://doi.org/10.1007/s11042-018-6784-9.
[59] N. E. Khalifa, M. Loey, and S. Mirjalili, "A comprehensive survey of recent trends in deep learning for digital images augmentation," *Artif. Intell. Rev.*, vol. 55, no. 3, pp. 2351–2377, 2022, https://doi.org/10.1007/s10462-021-10066-4.
[60] A. Bagaskara and M. Suryanegara, "Evaluation of VGG-16 and VGG-19 Deep Learning Architecture for Classifying Dementia People," *4th Int. Conf. Comput. Informatics Eng.*, pp. 1–4, 2021, https://doi.org/10.1109/IC2IE53219.2021.9649132.
[61] M. Shafiq and Z. Gu, "Deep Residual Learning for Image Recognition: A Survey," *Appl. Sci.*, vol. 12, no. 18, p. 8972, 2022, https://doi.org/10.3390/app12188972.
[62] S. Lu, B. Wang, H. Wang, L. Chen, M. Linjian, and X. Zhang, "A real-time object detection algorithm for video," *Comput. Electr. Eng.*, vol. 77, pp. 398–408, 2019, https://doi.org/10.1016/j.compeleceng.2019.05.009.
[63] J. Liu, R. Liu, K. Ren, X. Li, J. Xiang, and S. Qiu, "High-Performance Object Detection for Optical Remote Sensing Images with Lightweight Convolutional Neural Networks," *IEEE 22nd Int. Conf. High Perform. Comput. Commun. IEEE 18th Int. Conf. Smart City; IEEE 6th Int. Conf. Data Sci. Syst.*, pp. 585–592, 2020, https://doi.org/10.1109/HPCC-SmartCity-DSS50907.2020.00074.

## BIOGRAPHY OF AUTHORS

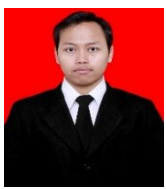
**Shoffan Saifullah** received a Bachelor's Degree in Informatics Engineering from Universitas Teknologi Yogyakarta, Indonesia, in 2015 and a Master's Degree in Computer Science from Universitas Ahmad Dahlan, Yogyakarta, Indonesia, in 2018. He is a lecturer at Universitas Pembangunan Nasional "Veteran" Yogyakarta, Indonesia. His research interests include image processing, computer vision, and artificial intelligence. He is currently a Ph.D. student at AGH University of Krakow, Poland, with a concentration in the field of artificial intelligence (bio-inspired algorithms), image processing, and medical analysis. Email: shoffans@upnyk.ac.id, and saifulla@agh.edu.pl. ORCID: https://orcid.org/0000-0001-6799-3834.

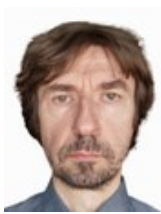
**Rafał Dreżewski** received the M.Sc., Ph.D., and D.Sc. (Habilitation) degrees in computer science from the AGH University of Krakow, Poland in 1998, 2005, and 2019, respectively. Since 2019, he has been an Associate Professor with the Institute of Computer Science, AGH University of Krakow, Poland. He is the author of more than 80 papers. His research interests include bio-inspired artificial intelligence algorithms and agent-based modeling and simulation of complex and emergent phenomena. Email: drezew@agh.edu.pl. ORCID: https://orcid.org/0000-0001-8607-3478.

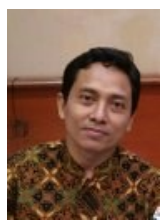
**Anton Yudhana,** B. Eng. and M. Eng. in electrical engineering from Institut Teknologi Sepuluh Nopember and Universitas Gajah Mada, Indonesia, in 2001 and 2005, respectively, and the Ph.D. degree from Universiti Teknologi Malaysia in 2011. He is an Associate Professor with the Electrical Engineering Department, Universitas Ahmad Dahlan, Indonesia, and the leader of Center for Electrical and Electronic Research and Development (CEERD). His current research interests include Agriculture Precision, Biomedic Engineering and Food Engineering Based on Internet of Things. Email: eyudhana@ee.uad.ac.id and eyudhana@mti.uad.ac.id.

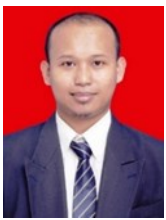
**Andri Pranolo** has a Master in Computer Science from Universitas Gadjah Mada, and currently Ph.D student in College Computer and Information, Hohai University, China. He is a lecturer at Informatics Department - Universitas Ahmad Dahlan (UAD) Indonesia, and member of Artificial Intelligent Research Group (AIRG) UAD with his focus on forecasting, and big data analytics. He active as an editorial board of some open access journals, such as International Journal of Advances in Intelligent Informatics. He is member of Institute of Electrical and Electronics Engineers (IEEE), and Association for Scientific Computing Electrical and Engineering (ASCEE). Email: andri.pranolo@tif.uad.ac.id. Orcid: https://orcid.org/0000-0002-3677-2788.

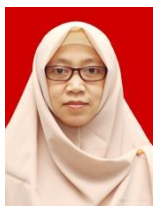

**Wilis Kaswidjanti** received a Bachelor's Degree in Computer Science from Universitas Gadjah Mada, Indonesia, in 1994 and a Master's Degree in Computer Science from Universitas Gadjah Mada, Yogyakarta, Indonesia, in 2002. She is a lecturer at Informatics Department - Universitas Pembangunan Nasional "Veteran" Yogyakarta, Indonesia. Her research interests include image processing, and artificial intelligence. Email: wilisk@upnyk.ac.id.

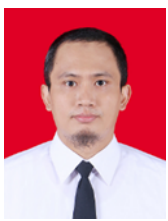

**Andiko Putro Suryotomo** is a lecturer at Department of Informatics, Universitas Pembangunan Nasional Veteran Yogyakarta, Indonesia. He obtained Bachelor's (S.Kom, 2010) and Master's (M.Cs., 2017) degree both from Department of Computer Science and Electronics, Universitas Gadjah Mada, Yogyakarta, Indonesia. His research interests include digital image processing, machine learning, and geoinformatics. E-mail: andiko.ps@upnyk.ac.id. ORCID: https://orcid.org/0000-0002-8070-6921.

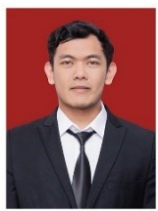

**Seno Aji Putra** was born in Bengkulu on December 6, 1999. He is a final year undergraduate student in Informatics from the Universitas Pembangunan Nasional "Veteran" Yogyakarta, Indonesia who took an interest in intelligent systems. In his final assignment, he focused on image processing, more precisely image processing of chicken egg fertility. Email: senoajiputra@gmail.com and 123180162@student.upnyk.ac.id. Orcid: https://orcid.org/0009-0003-1376-3525.

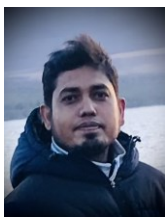

**Alin Khaliduzzaman** is a faculty member at Sylhet Agricultural University, Bangladesh. He obtained PhD in Bio-Sensing Engineering from Kyoto University, Japan. He has also served as a JSPS postdoctoral fellow at Kyoto University in Japan. During his PhD studies, he has opened up a new research domain in scientific world in the field of nondestructive chicken egg research and pioneered embryo grading concept for precision hatchery practices. He has been also awarded several conferences and Young Researcher's Awards due to his nondestructive egg and poultry research by various academic societies. He is currently working as a Research Assistant Professor at University of Hyogo, Japan. He edited the first and only the book "Informatics in Poultry Production" in this domain published by Springer in 2022. Email: khaliduzzamanfetsau2014@gmail.com. Orcid: https://orcid.org/0000-0001-6621-8430.

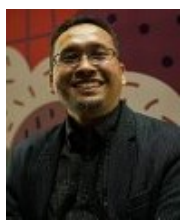

**Anton Satria Prabuwono** is Professor at Faculty of Computing and Information Technology in Rabigh, King Abdulaziz University. He holds M.Sc. in Computer Systems Engineering from University of East London and Ph.D. in Industrial Computing from National University of Malaysia (UKM). He started his academic career with Institute of Electronics, National Chiao Tung University, Taiwan and Faculty of Information and Communication Technology, University of Technical Malaysia Melaka (UTeM) in 2006 and 2007 respectively. He joined Faculty of Information Science and Technology, National University of Malaysia (UKM) in 2009. He then joined Faculty of Computing and Information Technology in Rabigh, King Abdulaziz University in 2013. He has received Best Publication Award from UTeM (2008), Excellent Service Award from UKM (2010), IEEE Gold Best Paper Award in IVIC 2011, Best Paper Award in ICEEI 2013, IEEE SMC Society Distinguished Lecturer 2016, and Outstanding Scientist Award 2016 from Venus International Foundation, India. He was an Erasmus Mundus Visiting Professor at Department of Mechanical Engineering and Mechatronics, Karlsruhe University of Applied Sciences, Germany. He was also Visiting Professor at Robotics

Lab, Japan Advanced Institute of Science and Technology. Before joining university, he has six years of experience in various multinational companies such as Bumi Kaya Steel, Samsung Electronics, and Coca-Cola Amatil. He was a Technical Service Manager at Coca-Cola Amatil Indonesia in 1997-2002. He has served as Editor, Technical Committee, and Reviewer in many International Journals and Conferences. He has published over 180 papers in Book Chapters, Journals and Proceedings. He is a member of ACM, senior member of IEEE Systems, Man and Cybernetics Society, and IEEE Robotics and Automation Society. In general, his research interests include machine vision, intelligent robotics, and autonomous systems. Email: aprabuwono@kau.edu.sa. Orcid: https://orcid.org/0000-0003-3337-6605.

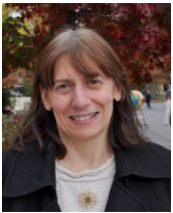

**Nathalie Japkowicz** is a Professor of Computer Science at American University. She was previously with the School of Electrical Engineering and Computer Science at the University of Ottawa where she lead the Laboratory for Research on Machine Learning for Defense and Security. Over the years, she has supervised over thirty graduate students, received funding from Canadian Federal and Provincial institutions (NSERC, DRDC, Health Canada, OCE, MITACS CITO), worked with private companies (Girih, Larus Technologies, Weather Telematics, TechInsights, Ciena) and published over 100 articles, papers and books including Evaluating Learning Algorithms: A Classification Perspective, with Mohak Shah (Cambridge University Press, 2011) and Big Data Analysis: New Algorithms for a New Society, with Jerzy Stefanowski (Springer, 2016). Email: nathalie.japkowicz@american.edu.